%% file: main.tex
\documentclass[11pt]{article}

\usepackage[final]{acl}

\renewcommand{\arraystretch}{0.95}

\usepackage{times}
\usepackage{latexsym}

\usepackage[T1]{fontenc}

\usepackage[utf8]{inputenc}
\usepackage{amsmath}
\usepackage{amssymb}
\usepackage{enumitem}
\usepackage{comment}
\usepackage{makecell}
\usepackage{booktabs}
\usepackage{multirow}
\usepackage{colortbl}
\usepackage{xcolor}
\usepackage{array}

\definecolor{ourblue}{RGB}{230, 240, 252}
\definecolor{ceilinggray}{RGB}{242, 242, 238}
\definecolor{gainfg}{RGB}{20, 110, 60}
\definecolor{lossfg}{RGB}{170, 40, 40}
\definecolor{frzrow}{RGB}{245, 245, 245}
\definecolor{prtrow}{RGB}{230, 243, 255}

\newcommand{\best}[1]{\textbf{#1}}
\newcommand{\gain}[1]{\textcolor{gainfg}{#1}}
\newcommand{\loss}[1]{\textcolor{lossfg}{#1}}

\newcommand{\Frz}{\textendash}    
\newcommand{\Prt}{$\bullet$}      
\usepackage[most]{tcolorbox}
\definecolor{promptblue}{RGB}{70, 130, 180}
\definecolor{promptgreen}{RGB}{60, 130, 100}

\newtcolorbox{promptbox}[2][promptblue]{
  enhanced,
  colback       = #1!4!white,
  colframe      = #1!80!black,
  colbacktitle  = #1!85!black,
  coltitle      = white,
  fonttitle     = \bfseries\small,
  title         = {#2},
  boxrule       = 0.6pt,
  titlerule     = 0pt,
  arc           = 3pt,
  left=8pt, right=8pt, top=6pt, bottom=6pt,
  fontupper     = \small,
}

\usepackage{microtype}

\usepackage{inconsolata}

\usepackage{graphicx}

\graphicspath{{Figs/}}

\title{ReMAP-PET: Beyond Visual Understanding - Learning \textbf{Re}gion-Guided \textbf{M}etabolic \textbf{A}lignment Semantics from Brain \textbf{P}ET}

\author{
Dasen Dai$^{1,}$\thanks{These authors contributed equally.},
Yanteng Zhang$^{2,}$\footnotemark[1]$^{,}$\thanks{Corresponding \& Project lead: \texttt{yntn32@outlook.com}},
Shuoqi Li$^{1,}$\footnotemark[1],
Yuxiang Wei$^{2}$,
Hongjie Yu$^{3}$, \\
Qingxin Zhang$^{4}$,
Qizhen Lan$^{5}$,
Jagath C. Rajapakse$^{6}$,
Vince D. Calhoun$^{2}$ \\
\\
$^{1}$ The Chinese University of Hong Kong, HKSAR \\
$^{2}$ TReNDS Center (Georgia State, Georgia Tech, Emory), Atlanta, USA \\
$^{3}$ ShanghaiTech University, Shanghai, P.R.China \\
$^{4}$ University of California, Berkeley, USA \\
$^{5}$ University of Texas Health Science Center at Houston, USA \\
$^{6}$ Nanyang Technological University, Singapore 
}

\begin{document}
\maketitle

\input{Sections/01_abstract}
\input{Sections/02_Intro_AI}
\input{Sections/03_related_work}
\input{Sections/04_methods}

\input{Sections/05_experiments}
\input{Sections/06_conclusion}

\input{Sections/07_acknowledgments}

\bibliography{reference} 
\newpage
\appendix
\input{Sections/08_appendix}

\end{document}

%% file: Sections/01_abstract.tex
\begin{abstract}

Positron Emission Tomography (PET) reveals brain metabolism and is clinically central to neurodegenerative disease assessment, yet existing 3D brain foundation models treat PET as generic volumetric data, missing the structured regional metabolic information that distinguishes it from structural neuroimaging. To address these limitations, we propose ReMAP-PET, a framework that moves beyond visual encoding by supervising a partially-tuned MedicalNet 3D ResNet-50 with brain regional standardized uptake value ratio (SUVR) profiles through joint regression and contrastive objectives, enabling the encoder to learn the metabolic semantics underlying PET modality. On 1015 paired PET--SUVR samples, ReMAP-PET achieves 0.070 SUVR MAE and 77.8\% PET$\rightarrow$SUVR Recall@1, substantially outperforming five frozen pretrained baselines. We further connect the metabolic embedding to clinical language via contrastive alignment with frozen BioClinicalBERT and demonstrate end-to-end PET-to-report generation through SUVR-constrained verbalization. Linear probing on diagnostic classification and cognitive regression tasks confirms that the embeddings retain clinically relevant information without task-specific fine-tuning. Our results show that grounding PET encoders in regional metabolic semantics---rather than treating PET as generic volumetric data---yields representations that are structured, interpretable, and language-compatible, pointing to a new direction for metabolic-aware PET understanding.

\end{abstract}

%% file: Sections/02_Intro_AI.tex
\section{Introduction}

Accurately modeling brain metabolic patterns is a core challenge in functional neuroimage analysis and neurodegenerative disease research, with important implications for understanding disease progression and supporting clinical diagnosis~\cite{perovnik2023functional}. As one of the clinical gold-standard functional modalities for dementia diagnosis, fluorodeoxyglucose (FDG)-PET reflects neuronal activity through glucose metabolism. Unlike structural MRI, which captures anatomical changes, FDG-PET reveals early functional abnormalities and cross-regional metabolic degeneration patterns, and has been widely shown to be closely associated with early screening, disease progression, and cognitive decline in Alzheimer’s disease (AD)~\cite{xie2024pet}. More importantly, PET possesses explicit quantitative medical properties: standardized uptake value ratios (SUVR) measured across brain regions provide physiologically meaningful descriptions of regional metabolism~\cite{teune2010typical}. For example, AD related hypometabolism typically first appears in the posterior cingulate cortex, precuneus, and temporoparietal regions. Compared with raw voxel intensities, these regional metabolic measurements exhibit substantially stronger structural organization and clinical interpretability~\cite{bailly2015precuneus, gunn2015quantitative}. Learning clinically interpretable metabolic representations from PET therefore matters directly for downstream cognitive disease analysis.

\begin{figure}[!ht]
    \centering
    \includegraphics[width=\columnwidth]{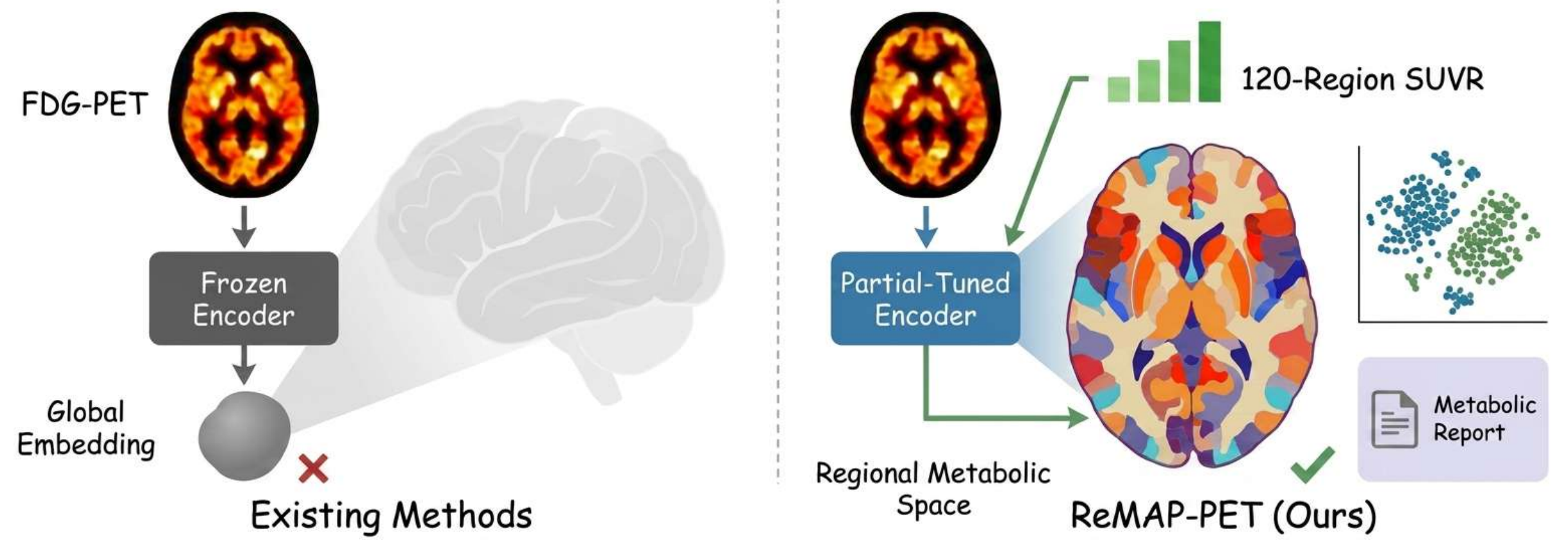}
    \caption{Existing methods treat PET as generic 3D volumes, extracting visual features that lack metabolic semantics (left). ReMAP-PET introduces region-level metabolic supervision to teach the encoder the physiological meaning behind PET imaging (right), producing representations that encode both visual and metabolic structure and enable cross-modal retrieval, clinical interpretation, and report generation.}
    \label{fig:intro}
\end{figure}

In recent years, the development of pretrained 3D medical image encoders has made brain representation learning increasingly feasible, enabling effective transfer to downstream neuroimaging tasks under limited annotation settings~\cite{yang2024deep,fang2025large}. However, nearly all existing frameworks are still built around structural MRI, with objectives primarily focused on visual texture modeling, global image representation learning, or anatomical structure reconstruction, while lacking explicit modeling of brain metabolic patterns themselves~\cite{zhou2025brain,falconnier2025representation}. As a result, when directly applied to PET scans, these methods still tend to treat PET as an ordinary 3D visual image without leveraging the metabolic information that PET inherently provides. This limits the model’s ability to capture the spatial organization of brain metabolism and to align representations with the structured clinical signals available in PET. A natural question is whether pretrained encoders can retain their general visual capability while also learning to encode the regional metabolic structure specific to PET. Rather than relying solely on image-level supervision, we aim to map PET into an interpretable metabolic space constrained by region-level measurements, so that the resulting representations capture neurodegenerative functional patterns and can support PET retrieval, automated report generation, and vision-language model (VLM) integration.

To address these challenges, we propose \textbf{ReMAP-PET} (\textbf{Re}gion-guided \textbf{M}etabolic \textbf{A}lignment with \textbf{P}artial-tuned \textbf{PET} Encoders), a method that grounds PET representation learning in regional metabolic activity. As illustrated in Figure~\ref{fig:intro}, ReMAP-PET uses 120-region SUVR profiles as structured supervision to reshape the embedding space of a pretrained MedicalNet 3D ResNet-50~\cite{chen2019med3d}, updating only the final residual stage (\texttt{layer4}) while preserving generic anatomical knowledge in the earlier layers. A joint objective combining SUVR regression and bidirectional PET--SUVR contrastive alignment produces embeddings that are both numerically predictive and semantically structured. We further connect this metabolic representation to clinical language: lightweight projection heads align the frozen PET embedding with region-text summaries encoded by frozen BioClinicalBERT~\cite{alsentzer2019publicly}, and an end-to-end report generation pipeline verbalizes the predicted SUVR profile into clinical-style metabolic summaries whose factual content is fully constrained by the underlying measurements.

Our experiments show three main results. First, regression alone produces accurate SUVR predictions but a degenerate embedding space; the contrastive term is what creates the structure needed for retrieval and downstream transfer. Second, partial tuning is architecture-dependent: the same last-block recipe that produces large gains on MedicalNet's ResNet yields negligible or negative effects on ViT-based and U-Net-based encoders, suggesting that ResNet's feature hierarchy provides a uniquely suitable bottleneck for metabolic adaptation. Third, linear probing on seven clinical tasks confirms that the metabolic embeddings carry clinically relevant information without task-specific fine-tuning, and the gains in metabolic alignment do not come at the expense of clinical signal.

Our contributions are as follows:
\begin{itemize}[nosep,leftmargin=*]
\item We propose a metabolic-aware representation learning paradigm for FDG-PET: instead of treating PET as generic volumetric data, we use 120-region SUVR profiles as structured metabolic supervision with joint regression and contrastive objectives, enabling the encoder to learn regional metabolic semantics beyond visual features.
\item We show that partial tuning of the final ResNet stage is uniquely effective for PET metabolic adaptation---the same recipe does not transfer to ViT or U-Net architectures---providing an empirical finding about when and why partial tuning works.
\item We connect the learned metabolic embedding to clinical language through PET--text contrastive alignment and SUVR-constrained report generation, completing an image-to-representation-to-language pipeline for FDG-PET.
\end{itemize}

%% file: Sections/03_related_work.tex
\section{Related Work}

\paragraph{3D brain foundation models.}
Pretrained 3D medical and brain encoders have become common in neuroimaging representation learning. MedicalNet provides a general 3D transfer-learning backbone, while BrainIAC~\citep{tak2026brainiac}, BrainFM~\citep{liu2025brainfm}, AnatCL~\citep{barbano2026anatcl}, and BrainSegFounder~\citep{cox2024brainsegfounder} explore different forms of brain-specific pretraining. Other volumetric backbones such as SAM-Med3D~\citep{wang2025sam} and SwinUNETR~\citep{hatamizadeh2021swin} are also widely used, with recent surveys summarizing this line of work. However, these models are largely designed for structural MRI, segmentation, or generic 3D representation learning, and do not explicitly model the regional metabolic patterns that are central to FDG-PET.

\paragraph{Medical vision-language alignment.}
Image--text contrastive learning has been widely used to align visual features with language~\citep{radford2021clip}. Medical variants such as ConVIRT~\citep{zhang2022contrastive}, MedCLIP~\citep{wang2022medclip}, PubMedCLIP~\citep{eslami2023pubmedclip}, and BiomedCLIP~\citep{zhang2023biomedclip} extend this idea to reports, captions, or biomedical text. Most of them focus on 2D modalities and use report- or label-level supervision. In the neuroimaging domain, recent efforts explore ROI-level multimodal alignment and cross-modality translation using structured regional features~\citep{xie2024cross}, though primarily for structural MRI rather than PET. For PET/CT reporting, region-aware models such as PETAR~\citep{maqbool2025petar} and Reg2RG~\citep{chen2025reg2rg} generate clinical text with explicit spatial grounding; however, these rely on free-form supervision and do not constrain report content to measured biomarkers. FDG-PET offers a different setting: language can be grounded in quantitatively measured regional metabolism, and the report content can be fully constrained by the SUVR measurement to avoid hallucination---a safety property that free-form supervision does not provide.

\paragraph{Parameter-efficient tuning.}
Full fine-tuning is often difficult in medical imaging because datasets are small and heterogeneous. Parameter-efficient methods, including adapters~\citep{houlsby2019parameter}, LoRA~\citep{hu2022lora}, prompt tuning~\citep{lester2021power}, and visual prompt tuning~\citep{jia2022visual}, adapt pretrained models while keeping most parameters fixed. Similar ideas have been used in medical image analysis, especially for adapting large segmentation models with limited annotations~\citep{zhang2023samed}. However, how partial tuning should be applied to 3D PET encoders remains less well studied.

\paragraph{Clinical semantics and structured biomarkers.}
Clinical text and structured biomarkers have received increasing attention in neuroimaging analysis. BERT~\citep{devlin2019bert} and BioClinicalBERT have shown the usefulness of pretrained language models for medical semantics. In brain PET, region-level measurements such as SUVR provide physiologically meaningful descriptions of metabolic status. Existing methods still mainly focus on image--text or image--label alignment, while joint modeling of 3D PET and structured metabolic profiles remains underexplored.

%% file: Sections/04_methods.tex
\section{Methods}
\label{sec:methods}

\begin{figure*}[!ht]
    \centering
    \includegraphics[width=1\linewidth]{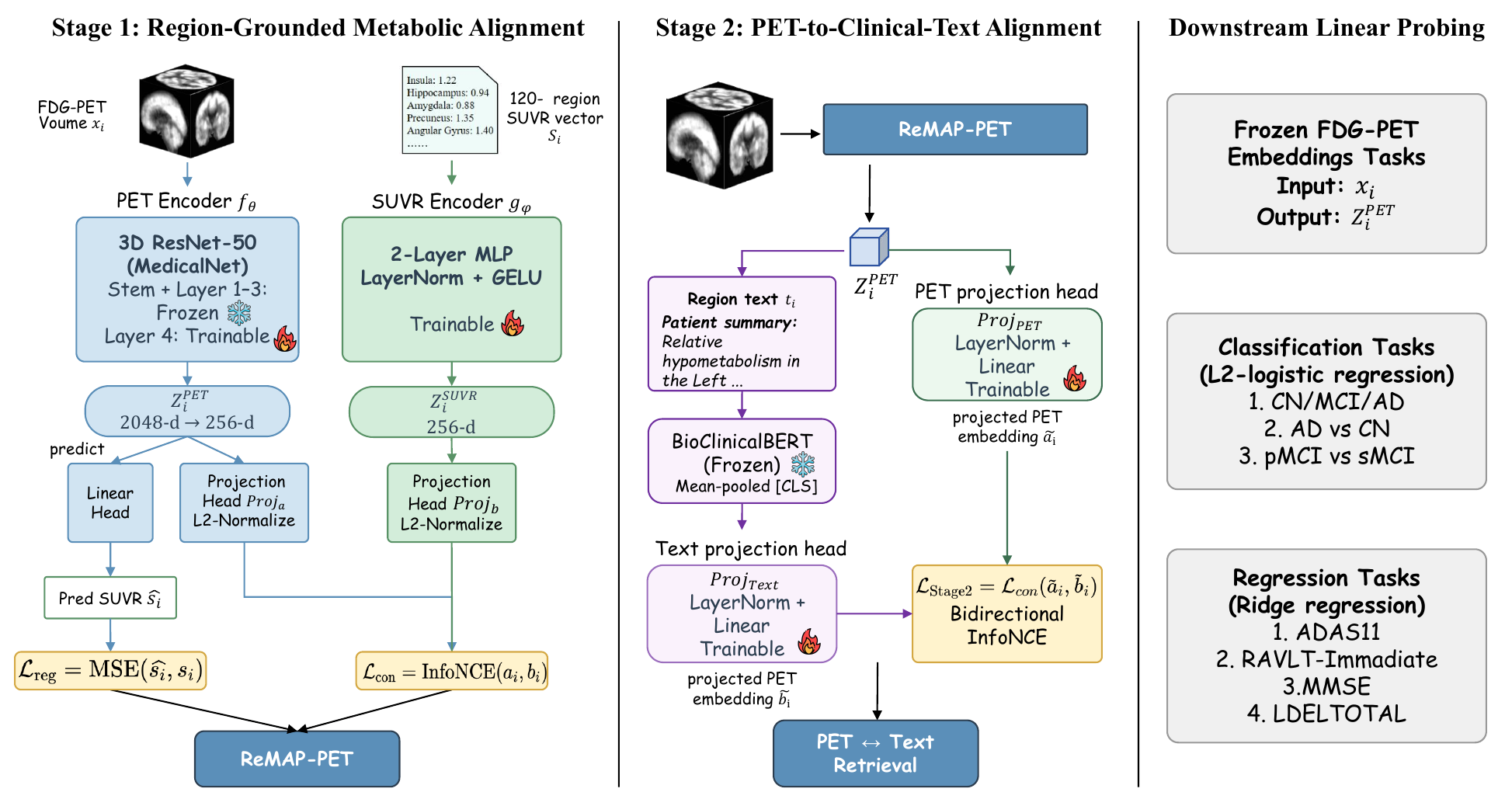}
    \caption{Overview of ReMAP-PET. \textbf{Stage~1} (left) aligns a partially-tuned 3D PET encoder with structured 120-region SUVR profiles via joint regression and contrastive objectives. \textbf{Stage~2} (center) connects the frozen metabolic embedding to clinical language through lightweight projection heads paired with frozen BioClinicalBERT. \textbf{Downstream probing} (right) evaluates the learned representations on diagnostic classification and cognitive regression tasks with no encoder fine-tuning.}
    \label{fig:method}
\end{figure*}
\subsection{Problem Setup}

As shown in Figure~\ref{fig:method}, the pipeline has two training stages followed by downstream probing. Each subject is a triplet \((x_i, s_i, t_i)\): a 3D FDG-PET volume \(x_i\), a 120-region SUVR vector \(s_i \in \mathbb{R}^{120}\) derived from the brain atlas~\cite{rolls2015implementation}, and a short text summary \(t_i\) deterministically generated from \(s_i\). In Stage~1, we train a PET encoder \(f_\theta\) jointly with a SUVR encoder \(g_\phi\) so that both modalities map into a shared embedding space; supervision comes from SUVR regression together with a cross-modal contrastive term. In Stage~2, \(f_\theta\) and a pretrained biomedical text encoder are both frozen, and only two small projection heads are trained to align the PET embedding with \(t_i\). We then evaluate the learned features by linear probing on clinical endpoints with no task-specific fine-tuning. Dataset statistics, splits, and preprocessing are given in Appendix~\ref{sec:appendix}.

\subsection{ReMAP-PET: Region-Grounded Metabolic Alignment}

The PET encoder in ReMAP-PET is a MedicalNet 3D ResNet-50. Rather than fine-tuning the whole network, we keep all early stages frozen and update only the last residual stage:
\begin{equation}
\theta = \{\theta_{\text{stem:layer3}}^{\text{frozen}},\; \theta_{\text{layer4}}^{\text{trainable}}\}.
\end{equation}
The intuition is straightforward: the early layers already encode generic anatomical structure from large-scale pretraining, and what is missing is a way to re-weight the highest-level features toward metabolic patterns specific to PET. We find in Section~\ref{sec:experiments} that this recipe is not universal. Applying the same protocol to ViT-based backbones (BrainIAC, SAM-Med3D) or to BrainFM's U-Net yields much smaller, and sometimes negative, gains. ResNet's last stage seems to act as a natural semantic bottleneck, which the transformer and U-Net backbones do not provide in the same way.

The SUVR vector is encoded by a small MLP \(g_\phi\) trained from scratch, giving \(z_i^{\text{suvr}} = g_\phi(s_i)\). On the PET side, a linear head \(h_\psi\) predicts the SUVR vector from the embedding, \(\hat{s}_i = h_\psi(z_i^{\text{pet}})\), and is supervised by mean squared error,
\begin{equation}
\mathcal{L}_{\text{reg}} = \frac{1}{B}\sum_{i=1}^{B} \| \hat{s}_i - s_i \|_2^2.
\end{equation}
The regression loss alone is enough to make the PET embedding numerically predictive of the SUVR profile, but it does not by itself produce an embedding space with a useful global structure. We therefore add a bidirectional contrastive term between normalized projections \(a_i, b_i\) of the PET and SUVR embeddings, with similarity \(\ell_{ij} = a_i^\top b_j / \tau\), where \(\tau\) is a learnable scalar initialized to \(0.07\) that controls the sharpness of the softmax distribution over negative pairs and is updated jointly with the model parameters via gradient descent:
\begin{equation}
\mathcal{L}_{\text{con}} = \tfrac{1}{2}\bigl[\operatorname{CE}(\ell, y) + \operatorname{CE}(\ell^\top, y)\bigr],\space  y_i = i.
\end{equation}
The full first-stage objective is \(\mathcal{L}_{\text{stage1}} = \lambda_{\text{reg}}\mathcal{L}_{\text{reg}} + \lambda_{\text{con}}\mathcal{L}_{\text{con}}\), with the two weights tuned on validation; we report the chosen values together with the ablation in Section~\ref{sec:experiments}.

To put ReMAP-PET in context, we compare it against five pretrained 3D encoders spanning ResNet, ViT, U-Net, and Swin architectures (Section~\ref{sec:exp-setup}). Each baseline is evaluated both fully frozen and with the same last-block partial-tuning recipe; the architecture-specific definitions of ``last block'' and the full comparison are reported in Section~\ref{sec:exp-stage1}.

\subsection{Linking PET to Clinical Text}

The second stage connects the metabolic embedding to natural language. Writing free-form radiology-style text for each scan would risk hallucinated findings, so we instead generate \(t_i\) from \(s_i\) by a fixed rule: we rank the 120 regions by SUVR, take the five lowest and five highest, map their identifiers to readable English names, and insert them into a short template that reports relative hypo- and hyper-metabolism. An optional LLM rewriting step is allowed only to improve fluency; it is explicitly forbidden to introduce regions, numbers, or diagnostic claims that are not already in the template. As a result, every factual statement in \(t_i\) is determined by the SUVR measurement. The template and rewriting prompt are reproduced in Appendix~\ref{sec:appendix}.

For alignment, we use BioClinicalBERT as the text encoder and take the mean-pooled last hidden state as the sentence representation. Both the text encoder and the first-stage PET encoder \(f_\theta\) are kept frozen, and we train only two small projection heads (LayerNorm followed by a linear layer),
\begin{equation}
\tilde{a}_i = \text{Proj}_{\text{pet}}(z_i^{\text{pet}}),\qquad
\tilde{b}_i = \text{Proj}_{\text{text}}(\bar{h}_i^{\text{text}}),
\end{equation}
using the same bidirectional contrastive loss as in the first stage. Because nothing inside the two encoders changes, any cross-modal alignment that emerges here can only come from the structure that the first stage has already put into the PET embedding space; this makes Stage~2 a fairly direct test of how well that structure transfers to language.

\subsection{Clinical Downstream Probing}

To assess whether the metabolic embeddings retain clinically relevant information, we freeze the Stage~1 encoder and fit linear probes on seven clinical endpoints: three diagnostic classification tasks (CN/MCI/AD, AD vs.\ CN, and pMCI vs.\ sMCI conversion) evaluated with logistic regression, and four cognitive score regressions (ADAS11, MMSE, RAVLT-Immediate, LDELTOTAL) evaluated with ridge regression. The regularization strength is selected on the validation split and each probe is evaluated once on the held-out test set; no encoder weights are updated. As an empirical ceiling, we apply the same probes to the ground-truth SUVR vectors. Hyperparameter grids and additional evaluation details are given in Appendix~\ref{app:evaluation}.

%% file: Sections/05_experiments.tex
\section{Experiments}
\label{sec:experiments}

\subsection{Setup}
\label{sec:exp-setup}

\input{Tables/table_main_stage1}

We work with an internal cohort of 1015 subjects, each providing a paired FDG-PET scan and 120-region SUVR profile, split by subject into 710 training, 152 validation, and 153 test cases. ADNI-derived clinical labels are matched to all subjects but are never seen during Stage~1 or Stage~2 training; they are used only for the linear probes in Section~\ref{sec:exp-clinical}. We compare ReMAP-PET against MedicalNet, BrainIAC, BrainFM, SAM-Med3D, and SwinUNETR, each evaluated fully frozen with only an MLP probe trained on top. For Stage~2, every PET encoder is paired with frozen BioClinicalBERT. Optimization settings, input resolutions, and the full list of metrics are given in Appendix~\ref{sec:appendix}; unless stated otherwise, all numbers are reported on the held-out test split.

\begin{figure}[!t]
    \centering
    \includegraphics[width=\columnwidth]{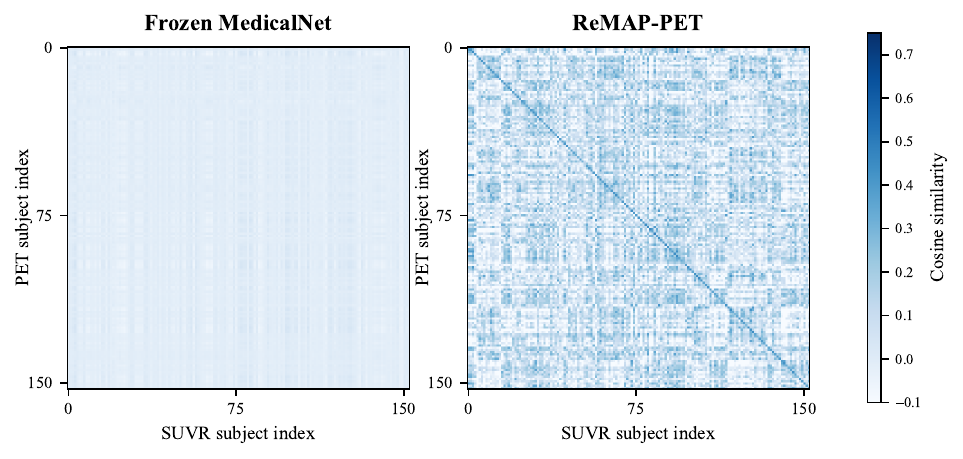}
    \caption{PET--SUVR cosine similarity matrices on the 153-subject test set. Left: Frozen MedicalNet produces near-uniform similarity (R@1 = 2.6\%). Right: ReMAP-PET yields a strong diagonal, indicating that each PET embedding is most similar to its paired SUVR embedding (R@1 = 77.8\%).}
    \label{fig:similarity}
\end{figure}

\begin{figure}[!t]
    \centering
    \includegraphics[width=\columnwidth]{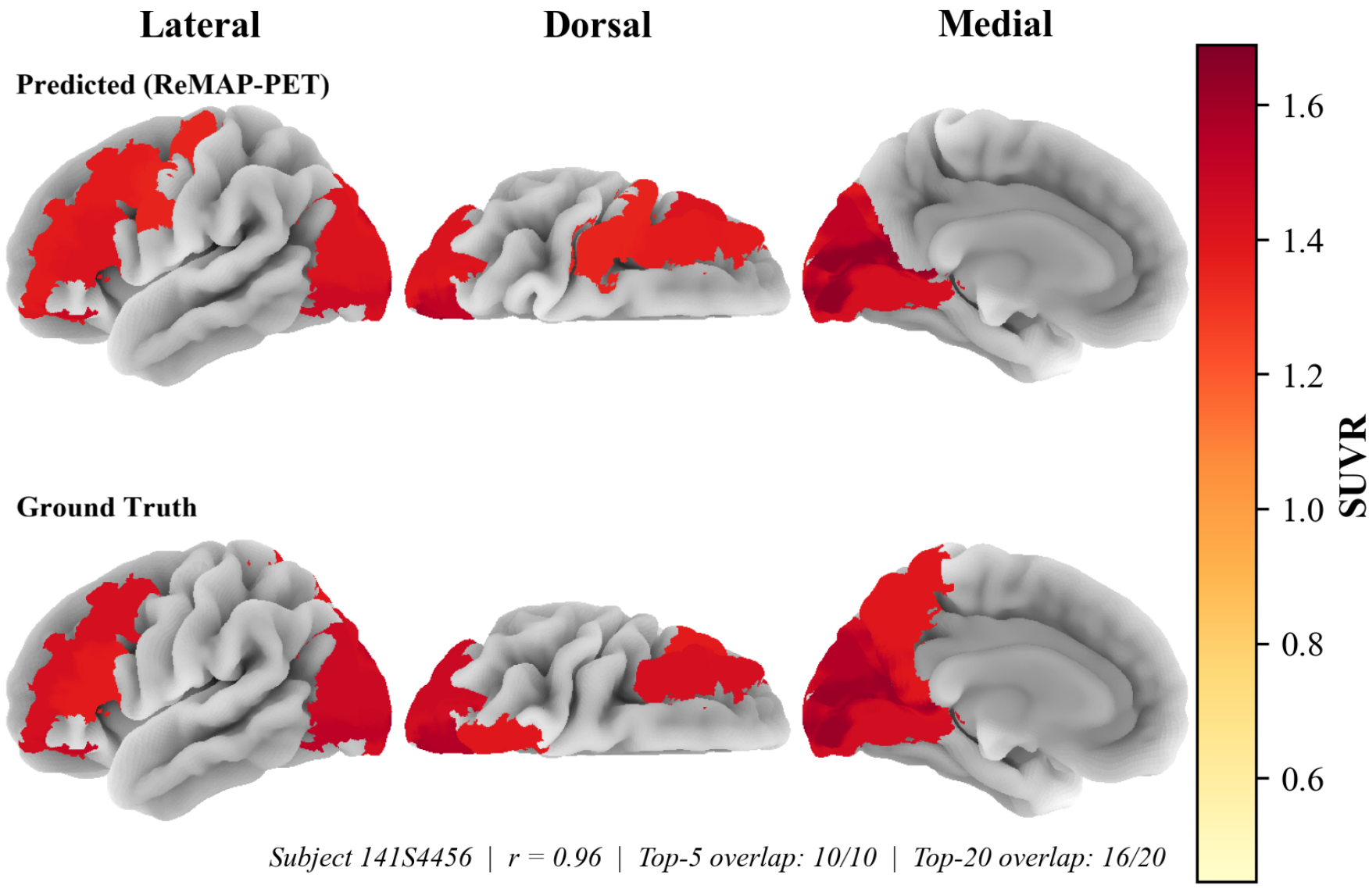}
    \caption{Predicted versus ground-truth 120-region SUVR profiles for a representative test subject. The close alignment between predicted (top) and true (bottom) SUVR values across the brain regions illustrates the quality of ReMAP-PET’s regional metabolic reconstruction (red denotes high metabolic activity).}
    \label{fig:predict_gt}
\end{figure}

\subsection{Stage~1: Aligning PET with Regional Metabolism}
\label{sec:exp-stage1}
Table~\ref{tab:main_stage1} compares ReMAP-PET against the five frozen baselines on the eight Stage~1 metrics. ReMAP-PET is the strongest method across the board. Against the best frozen baseline (BrainFM, a recent brain foundation model), it cuts SUVR MAE by roughly \(28\%\) (from \(0.097\) to \(0.070\)) and more than doubles PET\(\rightarrow\)SUVR Recall@1 (from \(0.37\) to \(0.78\)). Frozen MedicalNet, which shares the same backbone as ReMAP-PET, sits near the bottom of the table; its features alone clearly do not encode PET-specific metabolic semantics, and the gap to ReMAP-PET is entirely attributable to the partial-tuning recipe described in Section~\ref{sec:methods}. Figure~\ref{fig:similarity} visualizes this contrast: the frozen MedicalNet similarity matrix is near-uniform, whereas ReMAP-PET produces a strong diagonal indicating accurate cross-modal pairing. Figure~\ref{fig:predict_gt} further shows that the predicted SUVR profile closely matches the ground truth across the brain regions for a representative test subject.

\input{Tables/table_architecture}

A key question is whether these gains come from the partial-tuning recipe itself or from the specific backbone. Table~\ref{tab:architecture} applies the same last-block partial-tuning protocol to all five encoder architectures under identical training conditions. The recipe broadly helps MedicalNet's ResNet across all eight metrics, partially improves SAM-Med3D and SwinUNETR on retrieval, but degrades BrainFM on most metrics and barely affects BrainIAC. A plausible explanation is that ResNet's final bottleneck stage concentrates high-level semantic features in a single, clearly delineated block, making it a natural site for task-specific adaptation. By contrast, ViT self-attention blocks distribute information more uniformly across layers, and U-Net skip connections couple the deepest stage to the decoder, so tuning one block in isolation is less effective. This finding suggests that partial tuning is not a universal parameter-efficient alternative to full fine-tuning; its success depends on whether the architecture provides a clean semantic bottleneck for adaptation.

\subsection{Stage~2: Aligning PET with Clinical Text}
\label{sec:exp-stage2}

Stage~2 is designed as a diagnostic of the Stage~1 embedding: both the PET encoder (frozen after Stage~1) and BioClinicalBERT are fixed, so any retrieval signal must come from the metabolic structure that Stage~1 has learned. This means Stage~2 does not introduce new NLP capability; rather, it tests whether the regional patterns captured in Stage~1 are organized well enough to transfer to a language-grounded space through small projection heads alone. Table~\ref{tab:stage2} shows that ReMAP-PET achieves the best retrieval accuracy in both directions and the highest factual overlap between the retrieved and reference region summaries.

The absolute Recall@1 numbers look small, but the candidate pool contains 153 test subjects, so random chance is below \(0.7\%\). Across all encoders the Stage~2 ranking mirrors Stage~1, with BrainFM again the strongest baseline and BrainIAC the weakest. The high region-overlap scores (Low \(0.72\), High \(0.58\)) further suggest that even when ReMAP-PET does not retrieve the exact paired summary, the retrieved text describes a metabolically similar subject---a clinician receiving such a summary is getting actionable region-level information regardless of whether it comes from the exact paired subject, which is arguably the more clinically meaningful notion of correctness. Extended retrieval numbers (R@5) are reported in Appendix~\ref{app:stage2-full}.

\input{Tables/table_stage2}

\subsection{From Retrieval to Report Generation}
\label{sec:exp-report}

A natural follow-up to retrieval is whether the metabolic representation can also drive end-to-end report generation: given a PET scan alone, can we produce a metabolic summary that says the right thing? We test this by predicting the 120-region SUVR vector with ReMAP-PET, taking the five lowest- and five highest-metabolism regions, and passing them through Qwen3~\cite{qwen3} as a surface realizer; applying the same procedure to the ground-truth SUVR yields a per-subject reference paragraph.

Because both the predicted and reference reports are produced by the same verbalizer over a discrete region list, the only thing that varies between them is which regions the PET encoder identifies. These overlap with the reference at exactly the Top-5 high/low rates already reported for Stage~1 (Table~\ref{tab:main_stage1}), confirming that the verbalization step introduces no additional factual loss. Against ground-truth reference reports, the generated summaries achieve BLEU-1 of \(0.631\) and ROUGE-L of \(0.544\); residual surface differences reflect phrasing variation rather than factual disagreement. We view this as a read-out of the learned metabolic representation rather than a separate generation model. A per-subject prediction example and qualitative report comparison are provided in Appendix~\ref{app:case-study}.

\subsection{Clinical Probing}
\label{sec:exp-clinical}

Table~\ref{tab:clinical} reports linear probes on the three diagnostic classification tasks and on ADAS11---the cognitive score most directly tied to disease severity---with probes trained on the ground-truth 120-region SUVR vector as a non-trivial upper bound. The remaining three cognitive regression tasks (MMSE, RAVLT-Immediate, LDELTOTAL) are deferred to Appendix~\ref{app:clinical-full}.

ReMAP-PET achieves the highest AUROC on all three classification tasks: on the binary diagnostic extremes (AD vs.\ CN) it reaches \(0.946\), slightly above the SUVR ceiling (\(0.943\)). The gap is small (\(0.003\)) and within bootstrap uncertainty, but it is not contradictory: the SUVR ceiling reflects what a linear model can extract from 120 scalar values, whereas the 2048-dimensional PET embedding retains additional spatial and textural cues that the SUVR summary discards. ReMAP-PET is also the top encoder on the more challenging 3-way and pMCI vs.\ sMCI settings. We use linear probing rather than task-specific fine-tuning because our goal is to evaluate the intrinsic quality of the learned representation, not to maximize clinical performance; this is the standard protocol for representation evaluation in self-supervised learning. The regression numbers tell a more mixed story: no single encoder dominates and the SUVR ceiling is close to or slightly above every PET encoder. We read this as evidence that, once a backbone has reasonable 3D anatomical features, much of the clinically predictive signal is already accessible to a linear probe; the value added by ReMAP-PET is not primarily in raw clinical accuracy but in producing an embedding that is simultaneously good at SUVR prediction, retrieval, and downstream classification. The gains ReMAP-PET shows on Stage~1 metrics do not come at the cost of clinical signal---a failure mode we were initially concerned about.

\input{Tables/table_clinical}

\subsection{The Role of Contrastive Alignment}
\label{sec:exp-loss}

\begin{figure}[!t]
    \centering
    \includegraphics[width=\columnwidth]{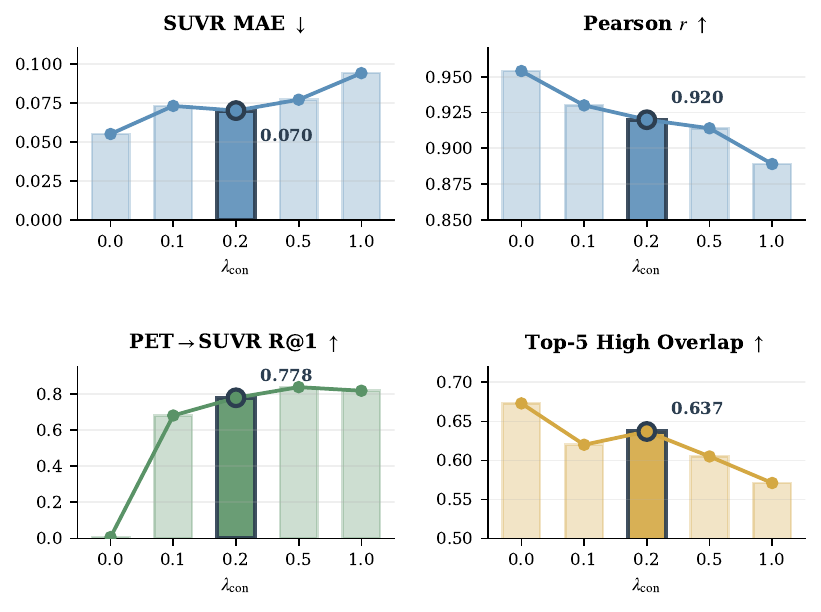}
    \caption{Contrastive weight ablation (\(\lambda_{\text{reg}}=1.0\) fixed, MedicalNet \texttt{layer4}). The highlighted bar marks the selected configuration (\(\lambda_{\text{con}}=0.2\)). Pure regression (\(\lambda_{\text{con}}=0\)) yields the lowest MAE but near-zero retrieval; adding contrastive alignment dramatically improves R@1 with modest regression cost.}
    \label{fig:ablation}
\end{figure}

Figure~\ref{fig:ablation} sweeps the contrastive weight \(\lambda_{\text{con}}\) with \(\lambda_{\text{reg}}=1.0\) fixed, using the MedicalNet \texttt{layer4} setup. The most informative comparison is at the two endpoints. With \(\lambda_{\text{con}}=0\) (pure regression), the model achieves its lowest SUVR error (MAE \(0.055\)) but its retrieval collapses to chance (R@1 \(\approx 0.007\), essentially \(1/153\)). The PET encoder has learned to predict SUVR values numerically, but its embedding space carries no useful global structure. Switching on the contrastive term at \(\lambda_{\text{con}}=0.1\) immediately raises R@1 to \(0.68\), and increasing it further trades a small amount of regression accuracy for additional retrieval quality.

We pick \(\lambda_{\text{con}}=0.2\) because it sits in the elbow of this tradeoff: regression MAE rises by less than a third compared to the pure-regression model, while R@1 reaches \(0.78\). More broadly, the ablation makes a point worth stating directly: regression alone is enough to fit SUVR values, but it is not enough to produce a representation that is useful for anything else. The contrastive term is what turns SUVR prediction into a structured metabolic embedding.

Beyond Recall@1, Table~\ref{tab:retrieval_detail} reports MRR and median rank for both retrieval directions. ReMAP-PET places the correct partner at rank~1 for the majority of subjects (MRR \(0.86\) PET\(\rightarrow\)SUVR, \(0.95\) SUVR\(\rightarrow\)PET; median rank~1 in both directions), while BrainFM frozen is the closest baseline (median rank~2) and BrainIAC barely aligns at all (median rank~75).

\input{Tables/table_retrieval_detail}

%% file: Tables/table_main_stage1.tex
\begin{table*}[!ht]
\centering
\small
\setlength{\tabcolsep}{4pt}
\renewcommand{\arraystretch}{1.15}
\begin{tabular}{l cc cc cc cc}
\toprule
\multirow{2}{*}{\textbf{Encoder}}
 & \multicolumn{2}{c}{\textit{SUVR Prediction}}
 & \multicolumn{2}{c}{\textit{Correlation}}
 & \multicolumn{2}{c}{\textit{Retrieval R@1}}
 & \multicolumn{2}{c}{\textit{Top-5 Region}} \\
\cmidrule(lr){2-3}\cmidrule(lr){4-5}\cmidrule(lr){6-7}\cmidrule(lr){8-9}
 & MAE\,$\downarrow$ & RMSE\,$\downarrow$
 & Pearson\,$\uparrow$ & Spearman\,$\uparrow$
 & P$\rightarrow$S\,$\uparrow$ & S$\rightarrow$P\,$\uparrow$
 & High\,$\uparrow$ & Low\,$\uparrow$ \\
\midrule
MedicalNet   & 0.117{\scriptsize$\pm$.009} & 0.153 & 0.739 & 0.845 & 0.026 & 0.046 & 0.429 & 0.729 \\
BrainIAC     & 0.125{\scriptsize$\pm$.009} & 0.161 & 0.707 & 0.815 & 0.013 & 0.020 & 0.343 & 0.680 \\
BrainFM      & \underline{0.097}{\scriptsize$\pm$.007} & \underline{0.126} & \underline{0.838} & \underline{0.885} & \underline{0.373} & \underline{0.458} & 0.471 & 0.736 \\
SAM-Med3D    & 0.115{\scriptsize$\pm$.009} & 0.147 & 0.829 & 0.858 & 0.163 & 0.196 & 0.463 & 0.724 \\
SwinUNETR    & 0.100{\scriptsize$\pm$.008} & 0.129 & 0.824 & 0.881 & 0.118 & 0.144 & \underline{0.507} & \underline{0.744} \\
\midrule
\rowcolor{ourblue}
\textbf{ReMAP-PET}
             & \best{0.070}{\scriptsize$\pm$.005} & \best{0.089} & \best{0.920} & \best{0.935}
             & \best{0.778} & \best{0.922} & \best{0.637} & \best{0.766} \\
\bottomrule
\end{tabular}
\caption{Stage~1 results on the held-out test set (153 subjects). Baselines use the named encoder fully frozen with only an MLP probe trained on top; ReMAP-PET uses MedicalNet 3D ResNet-50 with \texttt{layer4} unfrozen. P$\rightarrow$S / S$\rightarrow$P are PET-to-SUVR / SUVR-to-PET retrieval. Best in \best{}, second best underlined. MAE column shows bootstrap 95\% CI half-widths ($B{=}1000$).}
\label{tab:main_stage1}
\end{table*}

%% file: Tables/table_architecture.tex
\begin{table*}[!ht]
\centering
\small
\setlength{\tabcolsep}{4pt}
\renewcommand{\arraystretch}{1.2}
\begin{tabular}{l cc cc cc cc}
\toprule
\multirow{2}{*}{\textbf{Backbone}}
 & \multicolumn{2}{c}{\textit{SUVR Prediction}}
 & \multicolumn{2}{c}{\textit{Correlation}}
 & \multicolumn{2}{c}{\textit{Retrieval R@1}}
 & \multicolumn{2}{c}{\textit{Top-5 Region}} \\
\cmidrule(lr){2-3}\cmidrule(lr){4-5}\cmidrule(lr){6-7}\cmidrule(lr){8-9}
 & MAE\,$\downarrow$ & RMSE\,$\downarrow$
   & Pearson\,$\uparrow$ & Spearman\,$\uparrow$
   & P$\rightarrow$S\,$\uparrow$ & S$\rightarrow$P\,$\uparrow$
   & High\,$\uparrow$ & Low\,$\uparrow$ \\
\midrule
\rowcolor{ourblue}
\textbf{ReMAP-PET (ResNet)}
 & \best{0.070} & \best{0.089} & \best{0.920} & \best{0.935}
 & \best{0.778} & \best{0.922} & \best{0.637} & \best{0.766} \\
\midrule
\rowcolor{frzrow}
MedicalNet (ResNet) \Frz  & 0.117 & 0.153 & 0.739 & 0.845 & 0.026 & 0.046 & 0.429 & 0.729 \\
\rowcolor{frzrow}
BrainIAC (ViT) \Frz       & 0.125 & 0.161 & 0.707 & 0.815 & 0.013 & 0.020 & 0.343 & 0.680 \\
\rowcolor{prtrow}
BrainIAC (ViT) \Prt       & 0.119 & 0.155 & 0.731 & 0.847 & \loss{0.007} & 0.026 & 0.401 & 0.729 \\
\rowcolor{frzrow}
BrainFM (U-Net) \Frz      & 0.097 & 0.126 & 0.838 & \underline{0.885} & \underline{0.373} & \underline{0.458} & 0.471 & 0.736 \\
\rowcolor{prtrow}
BrainFM (U-Net) \Prt      & \loss{0.104} & \loss{0.134} & \loss{0.811} & \loss{0.843}
                     & \loss{0.261} & \loss{0.320} & 0.475 & \loss{0.682} \\
\rowcolor{frzrow}
SAM-Med3D (ViT) \Frz      & 0.115 & 0.147 & 0.829 & 0.858 & 0.163 & 0.196 & 0.463 & 0.724 \\
\rowcolor{prtrow}
SAM-Med3D (ViT) \Prt      & \underline{\gain{0.092}} & \underline{\gain{0.117}} & \underline{\gain{0.862}} & \gain{0.881}
                     & \gain{0.255} & \gain{0.346} & \underline{\gain{0.515}} & 0.731 \\
\rowcolor{frzrow}
SwinUNETR (Swin) \Frz     & 0.100 & 0.129 & 0.824 & 0.881 & 0.118 & 0.144 & 0.507 & \underline{0.744} \\
\rowcolor{prtrow}
SwinUNETR (Swin) \Prt     & 0.100 & 0.130 & 0.824 & 0.864 & \gain{0.235} & \gain{0.307} & \loss{0.418} & 0.726 \\
\bottomrule
\end{tabular}
\caption{Partial-tuning recipe applied across five encoder architectures. \colorbox{frzrow}{Gray}: frozen encoder + MLP probe; \colorbox{prtrow}{blue}: last-block partial tuning. \gain{Green} marks metrics that improve under partial tuning, \loss{red} marks degradation. The first row (MedicalNet frozen) is the starting point of ReMAP-PET.}
\label{tab:architecture}
\end{table*}

%% file: Tables/table_stage2.tex
\begin{table}[!t]
\centering
\small
\setlength{\tabcolsep}{4pt}
\renewcommand{\arraystretch}{1.15}
\begin{tabular}{l cc cc}
\toprule
\multirow{2}{*}{\textbf{Encoder}}
 & \multicolumn{2}{c}{\textit{R@1}}
 & \multicolumn{2}{c}{\textit{Region Overlap}} \\
\cmidrule(lr){2-3}\cmidrule(lr){4-5}
 & P$\rightarrow$T\,$\uparrow$ & T$\rightarrow$P\,$\uparrow$
 & Low\,$\uparrow$ & High\,$\uparrow$ \\
\midrule
MedicalNet   & 0.020 & 0.020 & 0.658 & 0.435 \\
BrainIAC     & 0.007 & 0.007 & 0.659 & 0.264 \\
BrainFM      & 0.026 & \underline{0.052} & \underline{0.698} & 0.460 \\
SAM-Med3D    & \underline{0.052} & 0.026 & 0.663 & \underline{0.465} \\
SwinUNETR    & 0.046 & \underline{0.052} & 0.694 & 0.459 \\
\midrule
\rowcolor{ourblue}
\textbf{ReMAP-PET}
             & \best{0.098} & \best{0.144} & \best{0.719} & \best{0.578} \\
\bottomrule
\end{tabular}
\caption{Stage~2 PET-text retrieval. All rows pair the named PET encoder (frozen) with frozen BioClinicalBERT. P$\rightarrow$T / T$\rightarrow$P are PET-to-text / text-to-PET retrieval; Region Overlap measures factual agreement between the query subject's reference summary and the summary retrieved for it. R@5 values are in Appendix~\ref{app:stage2-full}.}
\label{tab:stage2}
\end{table}

%% file: Tables/table_clinical.tex
\begin{table}[!t]
\centering
\resizebox{\columnwidth}{!}{%
\small
\setlength{\tabcolsep}{4pt}
\renewcommand{\arraystretch}{1.2}
\begin{tabular}{l ccc c}
\toprule
\multirow{2}{*}{\textbf{Encoder}}
 & \multicolumn{3}{c}{\textit{AUROC\,$\uparrow$}}
 & \textit{ADAS11} \\
\cmidrule(lr){2-4}\cmidrule(lr){5-5}
 & 3-way & AD/CN & pMCI/sMCI & MAE\,$\downarrow$ \\
\midrule
MedicalNet   & \underline{0.747}{\scriptsize$\pm$.060} & \underline{0.929}{\scriptsize$\pm$.069} & 0.719 & 3.990 \\
BrainIAC     & 0.652{\scriptsize$\pm$.064} & 0.809{\scriptsize$\pm$.114} & 0.660 & 4.598 \\
BrainFM      & 0.693{\scriptsize$\pm$.067} & 0.856{\scriptsize$\pm$.099} & 0.675 & 4.282 \\
SAM-Med3D    & 0.700{\scriptsize$\pm$.066} & 0.851{\scriptsize$\pm$.108} & \underline{0.759} & \underline{3.953} \\
SwinUNETR    & 0.684{\scriptsize$\pm$.072} & 0.912{\scriptsize$\pm$.078} & 0.737 & 4.072 \\
\midrule
\rowcolor{ourblue}
\textbf{ReMAP-PET}
             & \best{0.752}{\scriptsize$\pm$.062} & \best{0.946}{\scriptsize$\pm$.059} & \best{0.787} & \best{3.873} \\
\midrule
\rowcolor{ceilinggray}
\textit{SUVR (ceiling)}
             & \textit{0.774} & \textit{0.943} & \textit{0.835} & \textit{3.820} \\
\bottomrule
\end{tabular}%
}
\caption{Linear probing on three diagnostic tasks and ADAS11. The SUVR row uses the ground-truth 120-region SUVR as an upper bound. 3-way and AD/CN columns show bootstrap 95\% CI half-widths ($B{=}1000$). Full regression results are in Appendix~\ref{app:clinical-full}.}
\label{tab:clinical}
\end{table}

%% file: Tables/table_retrieval_detail.tex
\begin{table}[!t]
\centering
\small
\setlength{\tabcolsep}{4pt}
\renewcommand{\arraystretch}{1.2}
\begin{tabular}{l cc cc}
\toprule
\multirow{2}{*}{\textbf{Model}}
 & \multicolumn{2}{c}{\textit{P$\rightarrow$S}}
 & \multicolumn{2}{c}{\textit{S$\rightarrow$P}} \\
\cmidrule(lr){2-3}\cmidrule(lr){4-5}
 & MRR\,$\uparrow$ & MedR\,$\downarrow$
 & MRR\,$\uparrow$ & MedR\,$\downarrow$ \\
\midrule
\rowcolor{ourblue}
\textbf{ReMAP-PET}  & \best{0.857} & \best{1} & \best{0.949} & \best{1} \\
\midrule
\rowcolor{frzrow}
BrainFM              & \underline{0.533} & \underline{2}  & \underline{0.597} & \underline{2} \\
\rowcolor{prtrow}
BrainFM              & 0.446 & 3  & 0.480 & 3 \\
\rowcolor{frzrow}
SAM-Med3D            & 0.299 & 7  & 0.363 & 4 \\
\rowcolor{prtrow}
SAM-Med3D            & 0.446 & 3  & 0.499 & 3 \\
\rowcolor{frzrow}
SwinUNETR            & 0.244 & 9  & 0.280 & 7 \\
\rowcolor{prtrow}
SwinUNETR            & 0.382 & 5  & 0.462 & 3 \\
\rowcolor{prtrow}
BrainIAC             & \loss{0.037} & \loss{75} & \loss{0.067} & \loss{50} \\
\bottomrule
\end{tabular}
\caption{Extended PET--SUVR retrieval metrics. \colorbox{frzrow}{Gray}: frozen encoder; \colorbox{prtrow}{blue}: last-block partial tuning. MedR is the median rank of the correct cross-modal partner. \best{Bold}: best; \underline{underline}: second best.}
\label{tab:retrieval_detail}
\end{table}

%% file: Sections/06_conclusion.tex
\section{Conclusion}

We presented ReMAP-PET, an approach that grounds FDG-PET representation learning in regional metabolic semantics. By supervising only the final ResNet stage with 120-region SUVR profiles, the method learns embeddings that are predictive of regional metabolism, retrievable across modalities, and clinically useful under linear probing without task-specific fine-tuning. A comparison across five encoder families shows that the effectiveness of partial tuning depends on whether the backbone provides a clean semantic bottleneck, an observation that may carry over to other 3D medical adaptation settings. The learned metabolic structure transfers to a text-grounded space through frozen projection heads, and predicted SUVR profiles can be verbalized into reports whose content is fully determined by the measurement, sidestepping hallucination risks common in free-form medical text generation. We hope this work encourages further exploration of structured physiological supervision for building interpretable and language-compatible representations in functional neuroimaging. Future work should validate the approach on external cohorts with different tracers and scanners.

\paragraph{Limitations.}
Several limitations should be noted. First, all experiments use a single ADNI-derived cohort. Since PET acquisition requires the injection of radioactive tracers into participants, collecting PET data is considerably more complex and costly than acquiring conventional clinical imaging such as sMRI. As a result, publicly available brain imaging datasets containing PET scans are substantially smaller and less abundant worldwide. Under this practical constraint, this study first validated the proposed framework using the largest available collection of FDG-PET scans from the largest neuroimaging database, ADNI. Nevertheless, the generalizability of the model to other PET tracers, imaging protocols, or populations still requires further evaluation on external datasets. Second, the Stage~2 text descriptions are generated from a deterministic template rather than from free-form clinical reports, which limits the linguistic diversity of the training signal. Third, the partial-tuning finding is empirical---we provide a structural explanation for why ResNet benefits while ViT and U-Net do not, but a formal theoretical account is lacking. Finally, the current evaluation is restricted to linear probing; task-specific fine-tuning or integration into a full clinical decision pipeline may reveal different performance patterns.

%% file: Sections/07_acknowledgments.tex
\section*{Acknowledgments}
Data used in the preparation of this work were obtained from the ADNI. All participants provided informed consent, and ethical approval was obtained by the ADNI investigators from the respective institutional review boards. More information is available at \texttt{adni.loni.usc.edu}.

%% file: Sections/08_appendix.tex
\newpage
\section{Implementation Details}
\label{sec:appendix}

\subsection{Dataset and Preprocessing}

The dataset contains 1015 paired PET--SUVR samples derived from an internal ADNI-based cohort, including subjects with Alzheimer’s disease (AD), stable mild cognitive impairment (sMCI), progressive mild cognitive impairment (pMCI), and cognitively normal (CN) controls, split by subject into 710 training, 152 validation, and 153 held-out test cases. ADNI-derived clinical metadata are matched for all subjects and are used only for the linear probing experiments.

PET volumes are resampled to isotropic \(1.5\times1.5\times1.5\)~mm spacing, center-cropped or padded to \(96^3\) voxels, and intensity-normalized to zero mean and unit variance per volume. SAM-Med3D uses \(128^3\) inputs because of its learned positional embeddings. SUVR profiles are min--max normalized to \([0,1]\) per subject. The background region is excluded, leaving \(s_i \in \mathbb{R}^{120}\).

Per-subject min--max normalization removes absolute uptake differences caused by scanner variability, injection dose, and individual physiology, so that the model learns relative regional metabolic patterns rather than absolute tracer concentrations. All regression targets and prediction-error metrics (MAE, RMSE) are computed in this normalized [0,1] space. Figures that show SUVR profiles (e.g., Figure~\ref{fig:predict_gt}, Figure~\ref{fig:case_study}) display values mapped back to the original clinical scale for interpretability.

SUVR profiles are derived from the ADNI FDG-PET processing pipeline. Each PET volume is co-registered to the subject's MRI, spatially normalized to MNI space, and smoothed with an 8\,mm FWHM Gaussian kernel. Regional uptake is extracted using the AAL atlas~\cite{rolls2015implementation} (120 cortical and subcortical regions after excluding the background label). SUVR values are computed as the ratio of regional uptake to a cerebellar reference region. No partial-volume correction is applied; harmonization across scanners relies on the per-subject normalization described above.

\subsection{Backbones and Baselines}

The ReMAP-PET encoder uses MedicalNet 3D ResNet-50. The backbone consists of a stem block, four bottleneck stages, and global average pooling, yielding a 2048-dimensional feature vector before projection. ReMAP-PET keeps the stem through \texttt{layer3} frozen and updates only \texttt{layer4}. The SUVR encoder \(g_\phi\) is a two-layer MLP with LayerNorm and GELU.

We compare against four additional pretrained 3D encoders: BrainIAC, BrainFM, SAM-Med3D, and SwinUNETR. Each backbone is evaluated in two regimes: fully frozen with an MLP head, and last-block partial tuning using the same training recipe as ReMAP-PET.

\subsection{Optimization}

Stage~1 uses AdamW with learning rate \(1\times10^{-5}\), weight decay \(1\times10^{-4}\), batch size 4, and 50 epochs. The contrastive temperature \(\tau\) is initialized to 0.07. Unless otherwise stated, the loss weights are \(\lambda_{\text{reg}}=1.0\) and \(\lambda_{\text{con}}=0.2\), selected from the sweep in Table~\ref{tab:ablation_full}. Stage~2 uses AdamW with learning rate \(1\times10^{-4}\), batch size 16, and 20 epochs. All experiments use mixed-precision training on NVIDIA A40 GPUs.

\subsection{Region-Text Construction}

The controlled text summary used in Stage~2 is generated deterministically from the SUVR profile. For each subject, we rank the 120 regions by SUVR, select the five lowest- and five highest-metabolism regions, map atlas identifiers to readable English names, and insert them into the template:
\begin{quote}\small
FDG-PET regional metabolic summary. Relatively low metabolism is observed in [low-5 regions]. Relatively high uptake is observed in [high-5 regions].
\end{quote}

An optional Qwen3 rewriting step is used only as a surface realizer. The prompt instructs the model to preserve the given region names and polarity labels, and not to introduce new regions, measurements, or diagnostic claims.

\section{Evaluation Protocols}
\label{app:evaluation}

\subsection{Stage~1 and Stage~2 Metrics}

Stage~1 reports SUVR prediction error (MAE, RMSE), rank correlation between predicted and true SUVR profiles (Pearson \(r\), Spearman \(\rho\)), bidirectional PET--SUVR retrieval, and Top-5 high/low-metabolism region overlap.

Stage~2 reports bidirectional PET--text retrieval and retrieved-text region overlap. The main text reports Recall@1 and factual region overlap; full Recall@5 results are given in Appendix~\ref{app:stage2-full}.

\subsection{Clinical Probing}

For each Stage~1 encoder, PET embeddings are extracted with the encoder frozen. Classification probes use \(L_2\)-regularized logistic regression, with
\[
C \in \{0.01, 0.03, 0.1, 0.3, 1.0, 3.0, 10.0, 30.0\}.
\]
Regression probes use Ridge regression, with
\[
\alpha \in \{0.1, 0.3, 1.0, 3.0, 10.0, 30.0, 100.0\}.
\]
The regularization strength is selected on the validation split, and the selected probe is evaluated once on the test split. No encoder parameters are updated during probing.

Classification tasks are evaluated with AUROC in the main text; balanced accuracy and macro-F1 are also computed. Regression tasks are evaluated with MAE and Pearson \(r\), with Spearman \(\rho\) included in the appendix tables where useful.

\section{Additional PET--Text Retrieval Results}
\label{app:stage2-full}

Table~\ref{tab:stage2_full_appendix} gives the full Stage~2 retrieval results, including Recall@5. These numbers complement the compact single-column table in the main text.

\begin{table}[h]
\centering
\small
\setlength{\tabcolsep}{4pt}
\renewcommand{\arraystretch}{1.15}
\begin{tabular}{l cc cc cc}
\toprule
\multirow{2}{*}{Encoder}
 & \multicolumn{2}{c}{\textit{P\(\rightarrow\)T}}
 & \multicolumn{2}{c}{\textit{T\(\rightarrow\)P}}
 & \multicolumn{2}{c}{\textit{Overlap}} \\
\cmidrule(lr){2-3}\cmidrule(lr){4-5}\cmidrule(lr){6-7}
 & R@1 & R@5 & R@1 & R@5 & Low & High \\
\midrule
MedicalNet   & 0.020 & 0.144 & 0.020 & 0.092 & 0.658 & 0.435 \\
BrainIAC     & 0.007 & 0.039 & 0.007 & 0.046 & 0.659 & 0.264 \\
BrainFM      & 0.026 & 0.183 & 0.052 & 0.163 & 0.698 & 0.460 \\
SAM-Med3D    & 0.052 & 0.170 & 0.026 & 0.163 & 0.663 & 0.465 \\
SwinUNETR    & 0.046 & 0.196 & 0.052 & 0.144 & 0.694 & 0.459 \\
\midrule
\rowcolor{ourblue}
ReMAP-PET    & 0.098 & 0.392 & 0.144 & 0.340 & 0.719 & 0.578 \\
\bottomrule
\end{tabular}
\caption{Full Stage~2 PET--text retrieval results. All rows use frozen BioClinicalBERT on the text side; only the PET encoder changes. P\(\rightarrow\)T denotes PET-to-text retrieval and T\(\rightarrow\)P denotes text-to-PET retrieval.}
\label{tab:stage2_full_appendix}
\end{table}

\section{Additional Clinical Probing Results}
\label{app:clinical-full}

The main text reports the three diagnostic classification tasks and ADAS11. Table~\ref{tab:clinical_regression_appendix} reports the remaining cognitive regression tasks. The pattern is similar to ADAS11: ReMAP-PET is competitive, but no single encoder dominates every score.

\begin{table*}[t]
\centering
\small
\setlength{\tabcolsep}{4pt}
\renewcommand{\arraystretch}{1.15}
\begin{tabular}{l ccc ccc ccc}
\toprule
\multirow{2}{*}{Encoder}
 & \multicolumn{3}{c}{\textit{MMSE}}
 & \multicolumn{3}{c}{\textit{RAVLT Immediate}}
 & \multicolumn{3}{c}{\textit{LDELTOTAL}} \\
\cmidrule(lr){2-4}\cmidrule(lr){5-7}\cmidrule(lr){8-10}
 & MAE\(\downarrow\) & \(r\uparrow\) & \(\rho\uparrow\)
 & MAE\(\downarrow\) & \(r\uparrow\) & \(\rho\uparrow\)
 & MAE\(\downarrow\) & \(r\uparrow\) & \(\rho\uparrow\) \\
\midrule
MedicalNet   & \best{1.698} & 0.631 & 0.525 & 9.152 & 0.519 & 0.427 & \best{3.389} & 0.485 & 0.451 \\
BrainIAC     & 2.044 & 0.380 & 0.322 & 9.762 & 0.386 & 0.283 & 3.581 & 0.342 & 0.323 \\
BrainFM      & 1.969 & 0.506 & 0.410 & 9.170 & 0.472 & 0.379 & 3.562 & 0.384 & 0.370 \\
SAM-Med3D    & 1.850 & 0.531 & 0.414 & \best{9.013} & 0.520 & 0.381 & 3.613 & 0.421 & 0.358 \\
SwinUNETR    & 1.884 & 0.513 & 0.387 & 9.416 & 0.462 & 0.379 & 3.454 & 0.390 & 0.401 \\
\midrule
\rowcolor{ourblue}
ReMAP-PET    & 1.715 & 0.620 & 0.520 & 9.047 & 0.547 & 0.430 & 3.409 & 0.477 & 0.439 \\
\midrule
\rowcolor{ceilinggray}
\textit{SUVR} & \textit{1.721} & \textit{0.617} & \textit{0.554}
              & \textit{9.183} & \textit{0.534} & \textit{0.487}
              & \textit{3.338} & \textit{0.494} & \textit{0.484} \\
\bottomrule
\end{tabular}
\caption{Additional cognitive regression probing results. Bold marks the best PET encoder per MAE column; the SUVR row uses the ground-truth 120-region SUVR vector as an empirical ceiling.}
\label{tab:clinical_regression_appendix}
\end{table*}

\section{Additional Ablation and Retrieval Details}
\label{app:additional-ablation}

\subsection{Full Loss-Weight Sweep}

The main text uses a compact loss-weight table. Table~\ref{tab:ablation_full} includes the full set of Stage~1 metrics for the \(\lambda_{\text{con}}\) sweep.

\begin{table*}[!ht]
\centering
\small
\setlength{\tabcolsep}{5pt}
\renewcommand{\arraystretch}{1.15}
\begin{tabular}{l cc cc cc cc}
\toprule
\multirow{2}{*}{\(\lambda_{\text{con}}\)}
 & \multicolumn{2}{c}{\textit{SUVR Prediction}}
 & \multicolumn{2}{c}{\textit{Correlation}}
 & \multicolumn{2}{c}{\textit{Retrieval R@1}}
 & \multicolumn{2}{c}{\textit{Top-5 Region}} \\
\cmidrule(lr){2-3}\cmidrule(lr){4-5}\cmidrule(lr){6-7}\cmidrule(lr){8-9}
 & MAE\(\downarrow\) & RMSE\(\downarrow\)
 & Pearson\(\uparrow\) & Spearman\(\uparrow\)
 & P\(\rightarrow\)S\(\uparrow\) & S\(\rightarrow\)P\(\uparrow\)
 & High\(\uparrow\) & Low\(\uparrow\) \\
\midrule
0.0 (reg.\ only)
        & \best{0.055} & \best{0.071} & \best{0.954} & \best{0.954}
        & 0.007 & 0.007 & \best{0.673} & \best{0.800} \\
0.1     & 0.073 & 0.096 & 0.930 & 0.941 & 0.680 & 0.915 & 0.620 & 0.773 \\
\rowcolor{ourblue}
0.2     & 0.070 & 0.089 & 0.920 & 0.935 & 0.778 & 0.922 & 0.637 & 0.766 \\
0.5     & 0.077 & 0.098 & 0.914 & 0.934 & \best{0.837} & 0.915 & 0.605 & 0.753 \\
1.0     & 0.094 & 0.119 & 0.889 & 0.910 & 0.817 & \best{0.928} & 0.571 & 0.760 \\
\bottomrule
\end{tabular}
\caption{Full loss-weight ablation with \(\lambda_{\text{reg}}=1.0\) fixed. Pure regression yields the lowest SUVR error but gives near-random retrieval. The selected setting, \(\lambda_{\text{con}}=0.2\), balances regression accuracy and cross-modal structure.}
\label{tab:ablation_full}
\end{table*}

\begin{table*}[!ht]
\centering
\small
\setlength{\tabcolsep}{5pt}
\renewcommand{\arraystretch}{1.15}
\begin{tabular}{l ccc ccc}
\toprule
\multirow{2}{*}{Model}
 & \multicolumn{3}{c}{\textit{P\(\rightarrow\)S}}
 & \multicolumn{3}{c}{\textit{S\(\rightarrow\)P}} \\
\cmidrule(lr){2-4}\cmidrule(lr){5-7}
 & R@5\(\uparrow\) & MRR\(\uparrow\) & MedR\(\downarrow\)
 & R@5\(\uparrow\) & MRR\(\uparrow\) & MedR\(\downarrow\) \\
\midrule
\rowcolor{ourblue}
ReMAP-PET          & \best{0.980} & \best{0.857} & \best{1}
                   & \best{0.987} & \best{0.949} & \best{1} \\
\midrule
BrainFM \Frz       & 0.732 & 0.533 & 2  & 0.765 & 0.597 & 2 \\
BrainFM \Prt       & 0.654 & 0.446 & 3  & 0.706 & 0.480 & 3 \\
SAM-Med3D \Frz     & 0.418 & 0.299 & 7  & 0.562 & 0.363 & 4 \\
SAM-Med3D \Prt     & 0.706 & 0.446 & 3  & 0.680 & 0.499 & 3 \\
SwinUNETR \Frz     & 0.353 & 0.244 & 9  & 0.412 & 0.280 & 7 \\
SwinUNETR \Prt     & 0.536 & 0.382 & 5  & 0.634 & 0.462 & 3 \\
BrainIAC \Prt      & \loss{0.039} & \loss{0.037} & \loss{75}
                   & \loss{0.059} & \loss{0.067} & \loss{50} \\
\bottomrule
\end{tabular}
\caption{Extended PET--SUVR retrieval metrics. \Frz~denotes a frozen encoder with an MLP head; \Prt~denotes last-block partial tuning. MedR is the median 1-based rank of the correct cross-modal partner.}
\label{tab:retrieval_detail_appendix}
\end{table*}

\subsection{Extended PET--SUVR Retrieval Metrics}

Table~\ref{tab:retrieval_detail_appendix} reports the full retrieval statistics behind the compact retrieval discussion in the main text.

\section{Per-Subject SUVR Prediction Example}
\label{app:case-study}

\begin{figure}[h]
    \centering
    \includegraphics[width=\columnwidth]{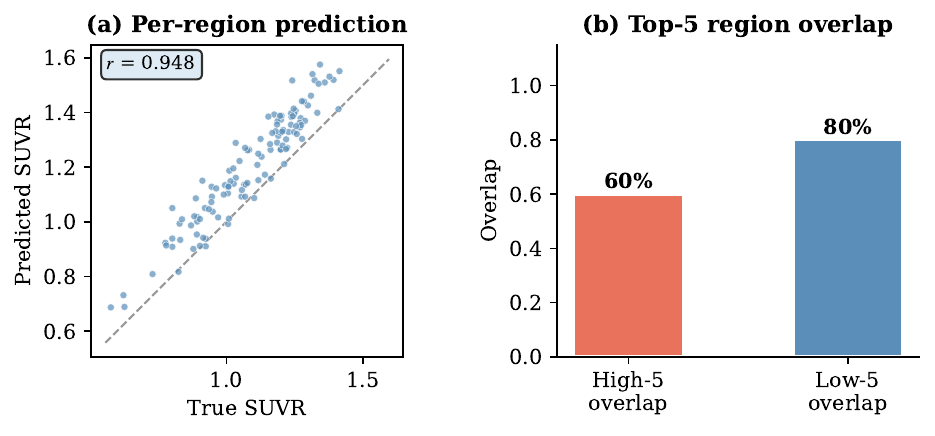}
    \caption{Per-subject SUVR prediction for a representative test subject (median Pearson $r$). (a) Predicted vs.\ true SUVR across 120 regions. (b) Top-5 high/low metabolism region overlap.}
    \label{fig:case_study}
\end{figure}

Figure~\ref{fig:case_study} shows a representative test subject (selected as the subject whose per-region Pearson $r$ is closest to the test-set median). The scatter plot confirms tight agreement between predicted and true SUVR across all 120 regions ($r = 0.948$). The Top-5 overlap bars show that ReMAP-PET correctly identifies 3 of 5 high-metabolism and 4 of 5 low-metabolism regions for this subject.

\section{Report Verbalization Examples}
\label{app:report-examples}

The two boxes below show verbalized reports for the representative test subject in Figure~\ref{fig:case_study}. Both are produced by the same Qwen3 prompt; they differ only in whether the input regions come from the predicted or ground-truth SUVR. The two reports agree on the main metabolic pattern (bilateral cerebellar and right temporal hypometabolism with frontoparietal preservation), with differences concentrated in nearby regions around the Top-5 boundary.

\begin{promptbox}[promptblue]{Generated Report (from ReMAP-PET Predicted SUVR)}
FDG-PET imaging demonstrates relatively reduced glucose metabolism involving the bilateral inferior cerebellar hemispheres and inferior vermis, as well as the right superior and middle aspects of the anterior temporal pole. In contrast, relatively preserved to increased radiotracer uptake is noted within the left lateral orbitofrontal cortex, left middle frontal gyrus, left angular gyrus, left posterior cingulate cortex, and right anterior orbitofrontal region. Overall, the examination demonstrates a heterogeneous metabolic pattern characterized by right temporopolar and bilateral inferior cerebellar hypometabolism with relative frontoparietal and cingulate preservation.
\end{promptbox}

\begin{promptbox}[promptgreen]{Reference Report (from Ground-Truth SUVR)}
FDG-PET imaging demonstrates relatively reduced glucose metabolism involving the bilateral cerebellar hemispheres, anterior cerebellar vermis, and superior right temporal pole. Conversely, relatively preserved to increased radiotracer uptake is observed within the left anterior and lateral orbitofrontal cortex, left middle frontal gyrus, left inferior parietal cortex, and left posterior cingulate cortex. Overall impression: The study demonstrates a heterogeneous metabolic pattern characterized by bilateral cerebellar and right temporal hypometabolism with relative left frontoparietal and posterior cingulate preservation.
\end{promptbox}